\pdfoutput=1
\documentclass[11pt]{article}

\usepackage[]{emnlp2022}

\usepackage{times}
\usepackage{latexsym}
\usepackage{amsmath}
\usepackage{amssymb}
\usepackage{slashbox}
\usepackage{enumitem}
\usepackage{float}
\usepackage[T1]{fontenc}
\usepackage[utf8]{inputenc}

\usepackage{microtype}

\usepackage{graphicx}
\usepackage{tabularx}
\title{\textit{FAST}: Improving Controllability for Text Generation with\\ \textit{F}eedback \textit{A}ware \textit{S}elf-\textit{T}raining}

\author{Junyi Chai, Reid Pryzant, Victor Ye Dong, Konstantin Golobokov, Chenguang Zhu, Yi Liu \\
        Microsoft Corporation \\
   \texttt{juchai,reidpryzant,victordong,kogolobo,chezhu,lewisliu@microsoft.com}}
\begin{document}
\maketitle
\begin{abstract}
Controllable text generation systems often leverage \emph{control codes} to direct various properties of the output like style and length. Inspired by recent work on causal inference for NLP, this paper reveals a previously overlooked flaw in these control code-based conditional text generation algorithms. Spurious correlations in the training data can lead models to incorrectly rely on parts of the input other than the control code for attribute selection, significantly undermining downstream generation quality and controllability. We demonstrate the severity of this issue with a series of case studies and then propose two simple techniques to reduce these correlations in training sets. The first technique is based on resampling the data according to an example's propensity towards each linguistic attribute (IPS). The second produces multiple counterfactual versions of each example and then uses an additional feedback mechanism to remove noisy examples (feedback aware self-training, FAST).  We evaluate on 3 tasks -- news headline, meta review, and search ads generation -- and demonstrate that FAST can significantly improve the controllability and language quality of generated outputs when compared to state-of-the-art controllable text generation approaches.
\end{abstract}

\section{Introduction}
In neural text generation, there is a growing interest in controlling the presence of particular linguistic attributes in the output text, for example sentiment, length, politeness, and topic \citep{Sennrich2016,Kikuchi2016,Ficler2017,Shen2021MReD}. This is typically accomplished via \textit{control codes}: categorical variables that represent the desired output property and are pre-pended to the model inputs during training and testing \cite{Keskar2019}. 

This paper builds on recent work in text-based causal inference \cite{Feder2021,Veitch2021,pryzant2020causal} to reveal a previously overlooked flaw in control code-based text generation systems: spurious correlations in the data can cause models to incorrectly rely on parts of the input \emph{other} than the control code for attribute selection, undermining downstream generation performance. 

For example, consider a system that generates news headlines while conditioning on article text and a control code for headline length (e.g. long for desktop, short for mobile) as in \citet{Murao2019}. We show in \S\ref{sec:spurious_correlation_case_study} that among publicly available news datasets, correlations exist between the contents of an article and the length of that article's title. Longer articles or articles about technical topics may be associated with longer titles. This leads NLP models to struggle at generating short titles from ``long title''-looking articles. 

\begin{figure}[t]
\centering
\includegraphics[width=\linewidth]{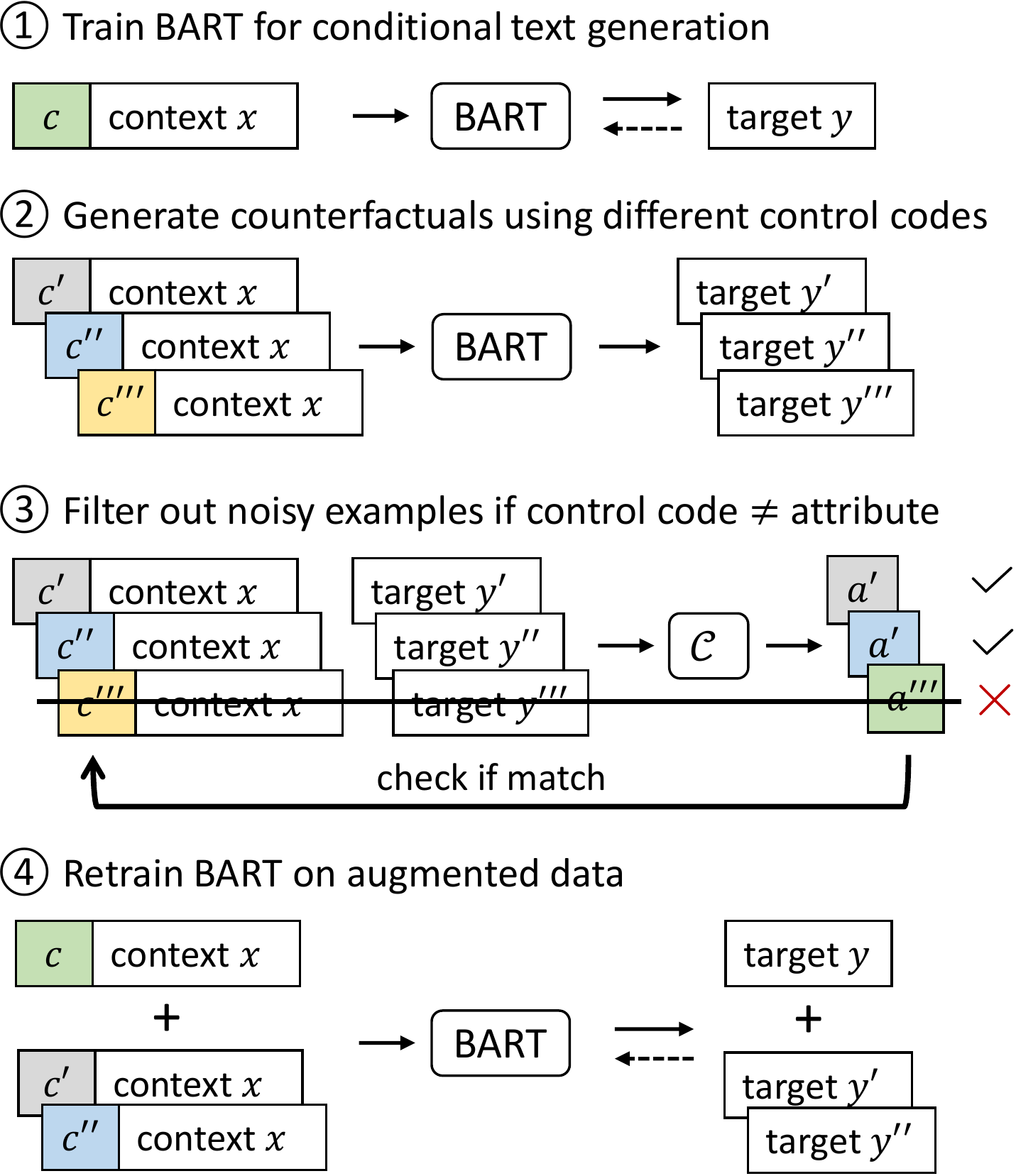}
\caption{Illustration of FAST algorithm.}
\label{fig:FAST_illustration}
\end{figure}

We show how this phenomenon can introduce confounding statistical relationships in the data, leading to assumption violations and significantly degrading model quality. Then we propose two simple data augmentation techniques for improving the issue. Both algorithms operate by breaking these spurious correlations and isolating the statistical relationship between control codes and linguistic attributes. In the first approach, we resample the training set according to an inverse propensity score (IPS, \citet{robins1994estimation}), boosting the presence of rare context-attribute combinations in the data. In the second approach (FAST) we train a preliminary model, use counterfactual data augmentation to generate all possible attributes for each example, then retrain on the counterfactually balanced dataset, as illustrated in Figure \ref{fig:FAST_illustration}. 

We conduct experiments in 3 conditional text generation scenarios: generating news headlines from article contents (controlling the headline lengths), generating the next sentence from preceding sentences (controlling the intent), and generating search ad copy from landing pages (controlling the rhetorical appeal of the ad). Our results suggest that FAST can significantly improve the controllability and language quality of state-of-the-art controllable generation systems. 

In summary, our contributions are:
\begin{itemize}[noitemsep]
    \item Identifying an important flaw with recent controllable text generation techniques and showing how this flaw can undermine model performance.
    \item A pair of simple yet effective data augmentation algorithms for dealing with this issue.
    \item Results and analysis demonstrating the efficacy of the proposed algorithms, and their ability to significantly improve controllability and language quality over state-of-the-art baselines.  
\end{itemize}

\section{Spuriously Correlated Control Codes}
\subsection{Controllable Generation}
We focus on the case of \emph{conditional} text generation, where the training data $D_{tr}=\{(x_1, y_1, a_1), ..., (x_n, y_n, a_n) \}$ is collection of triples consisting of context text $x$, output text $y$, and output linguistic attribute $a$. Note that in many practical scenarios, $a$ is inferred from $y$ by a classifier $\mathcal{C}(y)$, e.g. a rule or deep learning model. The goal is to learn a conditional language model (CLM) for $p(y \vert x, a)$, i.e. a text generation system which conditions on the context and attribute to generate texts that express the desired linguistic attribute.

In practice, the linguistic attributes $a$ are operationalized as control code tokens $c$ which are in one-to-one correspondence with the attributes (e.g. ``short'', ``long'') and pre-pended to the context $x$ before model input. This approach has been shown to be effective in both non-conditional \citep{Keskar2019,Ficler2017} and conditional \citep{Shen2021MReD,Fan2018} controllable text generation.   % which can simply be a rule  for determining the length category based on number of characters or a deep learning model for detecting an ad's category.

\subsection{Spurious Correlations}
\label{sec:spurious-corr}
In theory, the correspondence between the control code $c$ and linguistic attribute $a$ should cause models to rely on the control code to determine the linguistic properties of the generated output. This paper argues that in practice, parts of the context $x$ may be spuriously correlated with the attribute $a$, undermining the consistency and efficacy of control code-based systems.

These spurious correlations between the the contexts and attributes have a causal interpretation that explains how they can undermine model performance. The issue is that $p(a|x)\neq p(a)$, which is similar to a violation of the ignorability assumption in causal inference \citep{Feder2021}. This implies that any spurious correlations between the context $x$ and target attribute $a$ could represent backdoor paths that confound the model's learned relationship between the control code $c$ and the target attributes $a$. Thus, models are likely to depend on context \emph{beyond} the control code when determining output attributes, making them less likely to generalize to rare context/control-code combinations.

In this paper, we aim to break up these backdoor paths and prevent the model from learning spurious correlations. We accomplish this by modifying the training data in two ways such that $p(a|x)\approx p(a)$, with both techniques isolating the relationship between the control codes and target attributes.

\subsection{Inverse propensity score (IPS) resampling}\label{sec:IPS}

The first method we investigate for breaking the aforementioned spurious correlations leverages propensity scores. A propensity score is the conditional probability of an example being assigned to a treatment, given background variables \citep{ROSENBAUM1983}. It plays a central role in causal inference for dealing with spurious correlations in observational data and therefore it is a natural choice for us to try. In our case, the propensity score for the $i$th example is the conditional probability of the output text exhibiting linguistic attribute $a_i$ given the context $x_i$. This can be written as 
\begin{equation*}
w_i = p(a=a_i|x_i).
\label{eq:propensity_def}
\end{equation*}

Intuitively, examples with low propensity scores represent rare attribute-context combinations that are especially important to learn \citep{Tu2020}. Therefore, our procedure works by resampling the data with replacement, setting the sample weight of the $i$th example to $1 / w_i$. The procedure should work because the propensity scores of the resampled data should be close to uniform: $p(a_i \vert x_i) \propto w_i/w_i$.

In practice, we train a model to estimate propensity scores. For the experiments we finetune Roberta \citep{Liu2019} as a sequence classifier using $\{(x_1, a_1), ..., (x_n, a_n)\}$. We then use the model's probability prediction for the observed category $a_i$ as the propensity score estimate. We will refer to this estimator as $\mathcal{S}(a|x)$.

%\cz{better to put a few equations here to justify the uniform thing. And according to previous inequality, you assume p(a) is uniform?}

% Note that this procedure may excessively duplicate an example if its propensity score estimate is close to 0. Thus, we empirically cap the number of times an example can be resampled. 

% There are several drawbacks to this resampling approach. First, resampling may excessively duplicate an example if its propensity score estimate is close to 0. Thus, we empirically cap the number of times an example can be resampled. 
% Second, the propensity score is estimated by an imperfect model $\mathcal{S}$ so it can be noisy as well as biased. Third, it is possible that there are examples with wrong attribute labels, as the attributes are inferred by an imperfect classifier $\mathcal{C}$. Such examples are more likely to be considered rare and thus the noises get amplified after resampling.

%  A drawback from resampling is the duplication of examples, which can cause a language model to memorize rather than to generalize \citep{Lee2021,Carlini2021}.

\subsection{Feedback aware self-training (FAST)}\label{sec:fast}
The above IPS resampling procedure has several shortcomings, including the duplication of examples \citep{Lee2021,Carlini2021} and the noise/bias inherent to estimated propensity scores \cite{Pearl2009}. Therefore our second method, though originating with the same motivation and tackling the same issue, takes an orthogonal approach. First, use a separately trained model to produce multiple counterfactual target sequences for each context. Next, filter the data such that new target sequence expresses a different linguistic attribute. Then, we retrain on the new counterfactually balanced dataset. In detail, the steps are:
\begin{enumerate}[noitemsep]
  \item Train a conditional language model (CLM) using the standard control code approach on $D_{tr}$, which is denoted as CLM$_{baseline}$.
  \item Use CLM$_{baseline}$ to generate multiple outputs for each context $x_i$, one output for every control code except that which corresponds to the ground truth attribute. For example, the set of control codes used for datum $i$ would be $\{\forall c\in \{1,...,K\}, c \neq a_i\}$.
  \item Detect the linguistic attribute of the generation outputs with a classifier $\mathcal{C}$, and filter out examples where the predicted attribute does not match the inputted control code.
  \item Augment the original training set $D_{tr}$ with samples from Step 3 and retrain.
\end{enumerate}

Intuitively, this procedure should also drive the propensity scores of the data towards uniform and break the unwanted correlations between contexts and attributes, since every context becomes paired with multiple targets, each having a unique attribute. Step 3 uses feedback from the classifier $\mathcal{C}$ to remove noisy examples, preventing errors from propagating into the final model (\S\ref{sec:no_feedback_ablation}). We experiment with classifiers $\mathcal{C}$ that are given a prior,  trained on $(y, a)$ pairs from the training data $D_{tr}$, and trained on a separate dataset having similar properties.

% \cz{The descriptions look clear. Several general comments: 1. formalize CLM as a model such as L(y|x, a); 2. Name the different classifiers used in 3.1 and 3.2; 3. have an illustration for FAST; 4. in experiments investigate i) the acc about the attribute classifier using human eval; and ii) how many examples are filtered out in step 3, does it make data very imbalanced.}

\section{Experimental Setup}

We perform experiments in 3 important controllable generation settings: generating news headlines from article contents (controlling the headline lengths), generating the next sentence of a meta-review from preceding sentences and additional context (controlling the intent), and generating search ad copy from landing pages (controlling the rhetorical appeal of the ad). Our results suggest that the proposed methods can significantly improve the controllability and fluency of state-of-the-art baselines. 

\subsection{Datasets}\label{sec:dataset}

We experiment using 3 datasets (Table \ref{tbl:summary}) that reflect important real-world application scenarios for controllable generation systems.

\begin{table}[]
\centering
\resizebox{15em}{!}{
\begin{tabular}{lllll}
\hline
\multicolumn{5}{c}{PENS}                                                                                                               \\ \hline
\multicolumn{1}{l|}{Category} & \multicolumn{1}{l|}{train}  & \multicolumn{1}{l|}{dev}    & \multicolumn{1}{l|}{test rnd.} & test bal. \\ \hline
\multicolumn{1}{l|}{Short}    & \multicolumn{1}{l|}{31,245} & \multicolumn{1}{l|}{3,614}  & \multicolumn{1}{l|}{4,001}     & 5,509     \\
\multicolumn{1}{l|}{Long}     & \multicolumn{1}{l|}{57,351} & \multicolumn{1}{l|}{6,666}  & \multicolumn{1}{l|}{7,074}     & 5,509     \\
\multicolumn{1}{l|}{Total}    & \multicolumn{1}{l|}{88,596} & \multicolumn{1}{l|}{10,280} & \multicolumn{1}{l|}{10,240}    & 11,018    \\ \hline
\multicolumn{5}{c}{MReD} \\
\hline
\multicolumn{2}{l|}{Category}          & \multicolumn{1}{l|}{train} & \multicolumn{1}{l|}{dev}  & test      \\ \hline
\multicolumn{2}{l|}{Weakness}          & \multicolumn{1}{l|}{1,491}  & \multicolumn{1}{l|}{200}  & 200       \\
\multicolumn{2}{l|}{Strength}          & \multicolumn{1}{l|}{757}   & \multicolumn{1}{l|}{200}  & 200       \\
\multicolumn{2}{l|}{Decision}          & \multicolumn{1}{l|}{716}   & \multicolumn{1}{l|}{200}  & 200       \\
\multicolumn{2}{l|}{Rebuttal process}  & \multicolumn{1}{l|}{674}   & \multicolumn{1}{l|}{200}  & 200       \\
\multicolumn{2}{l|}{Abstract}          & \multicolumn{1}{l|}{581}   & \multicolumn{1}{l|}{200}  & 200       \\
\multicolumn{2}{l|}{Suggestion}        & \multicolumn{1}{l|}{438}   & \multicolumn{1}{l|}{200}  & 200       \\
\multicolumn{2}{l|}{Rating summary}    & \multicolumn{1}{l|}{338}   & \multicolumn{1}{l|}{159}  & 135       \\
\multicolumn{2}{l|}{Misc}              & \multicolumn{1}{l|}{225}   & \multicolumn{1}{l|}{143}  & 150       \\
\multicolumn{2}{l|}{AC disagreement}   & \multicolumn{1}{l|}{24}    & \multicolumn{1}{l|}{18}   & 18        \\
\multicolumn{2}{l|}{Total}             & \multicolumn{1}{l|}{5,244}  & \multicolumn{1}{l|}{1,520} & 1,503     \\
\hline
\multicolumn{5}{c}{Search Ads}                                                                                                         \\ \hline
\multicolumn{2}{l|}{Category}                               & \multicolumn{1}{l|}{train}  & \multicolumn{1}{l|}{dev}       & test \\ \hline
\multicolumn{2}{l|}{Product or Service} & \multicolumn{1}{l|}{1,771}  & \multicolumn{1}{l|}{44.6} & 43.1      \\
\multicolumn{2}{l|}{Call to action}     & \multicolumn{1}{l|}{1,207}  & \multicolumn{1}{l|}{37.5} & 36.6      \\
\multicolumn{2}{l|}{Location}           & \multicolumn{1}{l|}{931}   & \multicolumn{1}{l|}{22}   & 21.4      \\
\multicolumn{2}{l|}{Highlight}          & \multicolumn{1}{l|}{851}   & \multicolumn{1}{l|}{32}   & 30.8      \\
\multicolumn{2}{l|}{Inventory}          & \multicolumn{1}{l|}{590}   & \multicolumn{1}{l|}{19}   & 15.7      \\
\multicolumn{2}{l|}{Brand name}         & \multicolumn{1}{l|}{466}   & \multicolumn{1}{l|}{11.9} & 11        \\
\multicolumn{2}{l|}{Price}              & \multicolumn{1}{l|}{367}   & \multicolumn{1}{l|}{21.1} & 18.1      \\
\multicolumn{2}{l|}{Benefit}            & \multicolumn{1}{l|}{309}   & \multicolumn{1}{l|}{8.6}  & 8.6       \\
\multicolumn{2}{l|}{Customer problem}   & \multicolumn{1}{l|}{156}   & \multicolumn{1}{l|}{3.7}  & 3.9       \\
\multicolumn{2}{l|}{Total}              & \multicolumn{1}{l|}{6,649}  & \multicolumn{1}{l|}{200.5}  & 189.2  \\ \hline
\end{tabular}
}
\caption{Summary of PENS (top), MReD (middle) and Search Ads datasets (bottom, in thousands).}
\label{tbl:summary}
\end{table}

First, we use the \textbf{PENS} dataset released by Microsoft News \citep{Ao2021}. This task involves generating news headlines from news articles, while using a binary control code ``short'' or ``long'' to control the length of the generated headline (useful for mobile and desktop rendering). We use a length threshold of 55 to determine the long/short status of existing headlines in the data. We evaluate on these data using (1) random train/dev/test splits, and (2) a ``balanced'' test set. There are equal numbers of long and short headlines per article in this balanced test set. The headlines were sourced from 103 college students who wrote long or short headlines without seeing the original headlines, for an average of 3.7 headlines per article.

Second, we use the \textbf{MReD} dataset released by \citet{Shen2021MReD}. It consists of 4 years of ICLR meta reviews with each sentence being manually annotated into one of 9 categories. Using these data, our task is to follow the assisted writing scenario of \citet{Chen2019}. We generate the $i^{\mathrm{th}}$ sentence in the meta-review, controlling the intent of the generated sentence and conditioning on all preceding sentences and additional context (ratings, individual reviews). We reuse the original train/dev/test splits and randomly sample sentences with at least 4 words as the target sequences. For the training set, we pick one sentence per review. For dev and test sets, we pick multiple sentences per review while ensuring a nearly equal number of samples per category. To detect the categories of generated sentences, we train a Roberta-base classifier on 37,252 sentences (a superset of our generation training set), achieving a macro-F1 of 79\% on hold-out test set, implying that it has strong generalization capabilities.

Finally, we use a \textbf{Search Ads} dataset consisting of landing pages, search advertisements for those landing pages, and labels for those ads classifying them into one of 9 common advertising strategies. Here, the goal is to generate search ads (title and description) from landing pages while controlling the rhetorical appeal of the ad copy \citep{Golobokov2022}. To obtain the category labels, we apply a BERT-base-uncased model \citep{devlin-etal-2019-bert} trained on a separate dataset of 5,735 manually labeled ad-category pairs. This model achieves a macro-F1 score of 70\% on hold-out test set. Unlike the PENS data, the Search Ads data do not contain a balanced test set. However, the train, dev and test splits for the ads data contain an average of 1.9, 2.3 and 2.6 ads from different categories per landing page, respectively, so there is a moderate degree of category depth.

\subsection{Baselines}
We compare against five baselines: an uncontrolled system to establish a lower bound on performance, and four recently published neural controllable generation systems.

\paragraph{Uncontrolled} We train BART-base \citep{Lewis2020} for uncontrolled generation, where the model is only conditioned on the context.

\paragraph{BART+CTRL} We train BART-base for controllable generation using the standard control code approach \citep{Keskar2019}. The control code is represented as the name of the category (``long'', ``price'', etc). The paragraph symbol § is used as delimiter to separate control code and context. 

\paragraph{PPLM} We aim to enhance controllability of BART+CTRL by further steering its decoding towards the desired attribute. PPLM achieves this by using gradients from an attribute classifier $p(a|y)$ to update the CLM's hidden representations \citep{Dathathri2020}.

\paragraph{GeDi} This is a state-of-the-art technique for controlling open-ended and non-conditional generation \citep{Krause2021}. We adapt its weighted decoding formula to our \emph{conditional} generation setting by including a dependency on the context $x$:
\begin{equation}
p_w(y|x, c) \propto p(y|x)p(c|y)^\omega.
\label{eq:Gedi_eq1}
\end{equation}
The key insight from GeDi is to compute $p(c|y)$ using Bayes rule (i.e., leveraging $p(y|c)$). We train two BART-base models for $p(y|x)$ and $p(y|c)$ using the same procedure as the BART+CTRL baseline. We pick $\omega=4$ for PENS and MReD, and $\omega=3.5$ for Ads based on a brief hyperparameter search.

\paragraph{GeDi+x} Our last baseline involves further adapting GeDi to our application domain by conditioning everything on the context $x$ as well as the control code $c$, i.e. we concatenate the control code $c$ and the context $x$ when training BART+CTRL models. The new decoding formula is
\begin{equation}
p_w(y|x, c) \propto p(y|x, c)p(c|y)^\omega,
\label{eq:Gedi_x_eq1}
\end{equation}
and the Bayes approximation of $p(c|y)$ is
\begin{equation}
p(c|y) = \frac{p(c|x)p(y|x,c)}{\sum_{c'}p(c'|x)p(y|x,c')},\label{eq:Gedi_x_contrast_weight}
\end{equation}
where $p(c|x)$ is further dropped as it does not depend on $y$. We pick $\omega=1$ for PENS and $\omega=0.5$ for both MReD and Ads based on a brief hyperparameter search. Details of the above methods are in Appendix.

\subsection{Protocol}\label{sec:protocol}
Our implementation is largely based on Huggingface Transformers \citep{Wolf2020} except replacing beam search with a more efficient implementation \citep{fastseq}. We use \texttt{BART-base} and \texttt{Roberta-base} pretrained models to better emulate real-world production scenarios where smaller, more efficient models are favored. We use beam search with beam size 5 when decoding (except for PPLM, which uses greedy decoding). In all experiments, we train with 2 Nvidia V100 GPUs and inference with 1 GPU, both at fp16 precision.

For BART training, we optimize all models using Adamw \cite{loshchilov2017decoupled} and a learning rate of 1e-5 for PENS and 5e-5 for MReD and Ads datasets. We do not explicitly tune other hyperparameters. We train models, evaluating on the dev sets every epoch until the validation score begins to decrease. Then we pick the best-performing epoch based on ROUGE-1 with the dev set. All experiments are repeated with 5 random seeds when we report 95\% confidence intervals from a $t$-distribution. We consider $p<0.05$ to be statistically significant. More details are in the Appendix.

% For propensity score estimation, we finetune Roberta-base for sequence classification with the standard cross entropy loss. On PENS, we do a binary classification therefore use sigmoid to transform the score into probability. On Ads, we do a multi-class classification and therefore use softmax to convert scores into probabilities. We also use dev set to pick the best epoch during training based on AUC for PENS and accuracy for Ads. Learning rate is picked at 1e-5 with a brief hyperparameter tuning. More details are in the appendix.

\section{Experiments}

\subsection{Spurious Correlations}\label{sec:spurious_correlation_case_study}

We begin by empirically demonstrating the existence of spurious correlations that can degrade downstream model quality, and show how our algorithms reduce these correlations in the data. We show these trends via a series of case studies on the PENS news dataset. Similar studies on MReD and Search Ads datasets are in the Appendix.

\begin{table}[]
\centering
\resizebox{18em}{!}{
{%
\begin{tabular}{l|lll}
\hline
Method              & PENS & MReD & Ads \\
\hline
random guessing     & 50   & 11   & 11         \\ \hline
original       & 80   & 60   & 45         \\
IPS  & 52      &  18    & 11         \\
FAST & 59     &     15 & 24        \\ \hline
\end{tabular}%
}
}
\caption{Accuracy of predicting attribute of output from context on the original training set, as well as IPS resampled and FAST augmented training sets. }
\label{tbl:spurious_correlation}
\end{table}

\begin{table*}[]
\centering
\resizebox{\textwidth}{!}{%
{%
\begin{tabular}{l|llll|llll|llll}
\hline
             & \multicolumn{4}{c|}{PENS}                    & \multicolumn{4}{c|}{MReD}                 & \multicolumn{4}{c}{Search Ads}             \\
Method       & R1        & R2       & RL        & Acc      & R1       & R2      & RL       & Acc      & R1       & R2        & RL       & Acc      \\ \hline
Uncontrolled & 32.1±0.2  & 13.2±0.1 & 26.7±0.1  & --      & 18.7±0.5 & 4.1±0.2 & 16.1±0.4 & --      & 22.9±0.1 & 9.2±0.1   & 21.5±0.1 & --      \\
BART+CTRL    & \textbf{32.6}±0.1  & \textbf{13.4}±0.1 & \textbf{27.1}±0.1  & 78.0±0.7 & 21.4±0.2 & 5.6±0.2 & 18.4±0.2 & 76.5±1.9 & 27.8±0.1 & 11.9±0.1  & 26.2±0.1 & 68.1±1.2 \\
PPLM         & 29.5±0.1  & 10.8±0.1 & 24.4±0.1 & 76.3±0.6    &  \textbf{21.9}±0.2 &	5.1±0.2 & 18.4±0.1 & 74.7±1.1
   & 27.0±0.2 & 10.5±0.2 & 25.3±0.2 & 69.3±1.1 			
 \\ 
GeDi         & 31.7±0.1  & 12.6±0.1 & 26.2±0.1  & 78.5±0.6 & 17.3±0.6 & 3.7±0.4 & 15.3±0.5 & 74.9±5.3 & 23.2±0.6 & 8.2±0.4   & 21.9±0.5 & \textbf{83.3}±4.6 \\
GeDi+x       & 32.5±0.1  & 13.3±0.1 & 27.0±0.1  & \textbf{82.7}±0.6 & 19.9±0.4 & 4.9±0.2 & 17.3±0.3 & 83.7±0.6 & 27.7±0.2 & 11.8±0.1  & 26.1±0.2 & 77.9±0.9 \\ \hline
IPS          & 32.3±0.03 & 13.2±0.1 & 26.9±0.04 & 79.0±0.5 & 21.1±0.4 & 5.2±0.3 & 17.8±0.4 & 71.2±6.8 & 27.4±0.1 & 11.6±0.04 & 25.8±0.1 & 70.1±1.4 \\
FAST         & \textbf{32.5}±0.1  & \textbf{13.4}±0.1 & \textbf{27.1}±0.1  & \textbf{82.5}±0.5 & \textbf{21.9}±0.3 & \textbf{6.1}±0.2 & \textbf{18.9}±0.3 & \textbf{87.1}±0.7 & \textbf{28.1}±0.1 & \textbf{12.3}±0.1  & \textbf{26.5}±0.1 & \textbf{80.5}±0.2\\ \hline
\hspace{2mm}weak classifier& -- & -- & -- & -- & \textbf{21.8}±0.1 & \textbf{6.0}±0.1 & \textbf{18.9}±0.1 & 86.1±0.5 & \textbf{28.0}±0.2 & \textbf{12.1}±0.1 & \textbf{26.4}±0.1 & 76.6±0.7 \\ \hline
\end{tabular}%
}
}
\caption{Comparing different methods on PENS, MReD and Search Ads. We use ROUGE (R1, R2, and RL) to evaluate decoding quality on (1) the ``balanced'' PENS test set, (2) the default MReD test set and (3) the default Ads test set. We use Acc to evaluate controllability. This metric is calculated by generating using every control code, detecting the attribute category from the generated texts with task-specific classifiers or rules, then comparing the detected category with the control code to compute an accuracy. The best score is boldfaced. Multiple scores are boldfaced if there is no statistically significant difference. The last row is an ablation experiment for FAST using a weak classifier (logistic regression) in the feedback step.}
\label{tbl:main_result}
\end{table*}

In Section \ref{sec:spurious-corr}, we defined the spurious correlation issue as unwanted dependencies between the context and attribute: $p(a \vert x) \neq p(a)$. To reveal this property in the PENS dataset we finetune Roberta-base with a binary classification head to predict the attribute (long or short) from the context (news article), which also serves as $\mathcal{S}(a|x)$ in \S\ref{sec:IPS}. The model achieved an accuracy of 73\% on a hold-out test set, far better than random guessing (50\%) and the majority class (64\%), empirically confirming that $p(a \vert x) \neq p(a)$ and the context $x$ is strongly correlated with attributes $a$.

Next, we identify two sources of spurious correlation in the data. First, the length of a news article is positively correlated with its long/short headline status (point-biserial correlation $r_{pb}$ = 0.1, $p < 0.01$). Second, we find that certain words and phrases can be inappropriately correlated with the attributes. We train an l2-regularized logistic regression on the same task and data using bag-of-words features, then examine the features having the highest weights. The features most indicative of short headlines include niche topics like ``petfinder'', ``cartoonist's homepage'', and ``saildrone'' (a weather service) while words from established outlets that cover more general topics (``usa today'', ``cbsnewyork'') are associated with long headlines.

To show how spurious correlations can undermine downstream controllable generation performance, we train a BART-base model on the PENS dataset using the standard control code approach (BART+CTRL baseline). Next, for each article in the randomly split test set, we generate using all possible control codes (long, short) and score the outputs according to whether they are truly long or short. The system successfully generated the intended headline 89.6±0.6\% of the time in factual cases (when the control code matched the ground-truth attribute), but 64.1\%±0.6\%
 of the time in counterfactual cases. This suggests that models learn to rely on spurious correlations in the data, and this reliance can undermine generalization.

We proceed to show how our data augmentation algorithms produce datasets where these spurious correlations have been reduced in the data. We test the classifier predicting attributes from contexts on the original training set, as well as on the IPS resampled and FAST augmented training sets (Table \ref{tbl:spurious_correlation}). We find that the accuracy from the classifier is reduced greatly on the new training sets, implying that predictions become closer to random guessing as the spurious correlation is reduced and therefore $p(a|x) \rightarrow p(a)$ in the augmented data.

\subsection{Overall Generation Results}
We proceed to evaluate the impact of our algorithms on downstream controllable generation performance.

\begin{table*}[t]
\centering
\resizebox{34em}{!}{%
\begin{tabular}{l|lllll|l}
\hline
Method   & Language & Human-like & Factuality & Relevance & Overall  & Acc      \\ \hline
BART+CTRL & 93.5±0.7 & 99.3±0.2   & 99.4±0.2   & 99.7±0.2  & 92.8±0.7 & 52.1±2.3 \\
IPS      & 92.4±0.8 & 99.1±0.3   & 99.4±0.2   & 99.6±0.2  & 91.7±0.8 & 51.7±2.3 \\
FAST     & \textbf{94.9}±0.6 & \textbf{99.6}±0.2   & 99.2±0.3   & 99.7±0.2  & \textbf{94.1}±0.7 & \textbf{58.4}±2.3 \\ \hline
\end{tabular}%
}
\caption{Human evaluation results for Search Ads generation. The quality of each aspect (e.g., language) is measured by the percentage of samples in the good level. The overall quality is the percentage of samples with all 4 aspects in the good level. Controllability is measured by the accuracy between human labeled categories of the ads vs the control codes used during their generation. The best scores with statistical significance (under $Z$-test) is boldfaced.}
\label{tbl:Ads_human_eval}
\end{table*}

Table \ref{tbl:main_result} shows results using automatic evaluation metrics on the PENS, MReD and Search Ads datasets. One might hypothesize that FAST may outperform IPS resampling because 1) FAST does not create duplicate examples, and 2) there is no need to estimate propensity scores directly. Our results support this, as the proposed FAST algorithm always outperforms all baselines in either language quality or controllability or both, especially on MReD and Ads datasets. Among the 3 baselines, GeDi has much lower ROUGE, possibly because it only combines context and control code at decoding time, while the other methods encode context and control code together. GeDi+x improves controllability over BART+CTRL significantly, but slightly decreases ROUGE. FAST improves controllability over BART+CTRL to a similar degree, while maintaining ROUGE on PENS and even improving ROUGE on MReD and Ads. On the other hand, IPS improves controllability slightly over BART+CTRL on two datasets PENS and Ads, which suggests that IPS resampling is helpful in preventing the model from learning the spurious correlation. However, it hurts ROUGE on all datasets, which is likely due to the duplication of data (\S\ref{sec:IPS_sub_sampling}). It appears surprising that PPLM can have lower controllability than BART+CTRL even though it applies additional steering during decoding. This is because PPLM uses greedy decoding. For example on MReD, switching from beam search to greedy decoding, the control accuracy of BART+CTRL decreases from 76.5\% to 70\%. PPLM improves it to 74.7\%, but it is still behind BART+CTRL with beam search.

We proceed to conduct a human evaluation of downstream generation quality for the Search Ads dataset (Table \ref{tbl:Ads_human_eval}). We compare our IPS and FAST methods against BART+CTRL, omitting the PPLM and GeDi baselines because they are prohibitively expensive for many real-world applications. 

We use the models to generate ad copy in all 9 categories for each landing page, then present each generation to a panel of five professional judges. The judges evaluated each example in 4 aspects: whether the text is grammatically fluent (``language quality''), whether it was human-like and realistic, whether it was factual, and whether it is relevant to the landing page. Each aspect was rated according to a binary good/bad scale, then we report the average rating. Judges also categorized the generation into one of the 9 attribute categories summarized in Table \ref{tbl:summary}, which we converted into a measure of model controllability by calculating the accuracy between the input control codes and human labeled output categories. More details can be found in the Appendix.

The human evaluation results are consistent with automatic evaluation metrics. The proposed FAST outperforms the BART+CTRL baseline in both language quality and controllability, whereas the proposed IPS algorithm underperforms its baseline in terms of language quality while there is no perceptible difference in controllability.  %For title, FAST is better than baseline with statistically significance at $p<0.05$. For description, IPS is worse than baseline with statistically significance at $p<0.01$, while FAST seems to be better than baseline, though at a mild significance level $p=0.09$. In the human-like aspect, FAST is better than baseline with statistically significance at $p<0.05$. For controllability, FAST significantly outperforms baseline with about 6-7\% improvement in the accuracy of controlling generation attribute.

\begin{figure}[]
\includegraphics[width=20em]{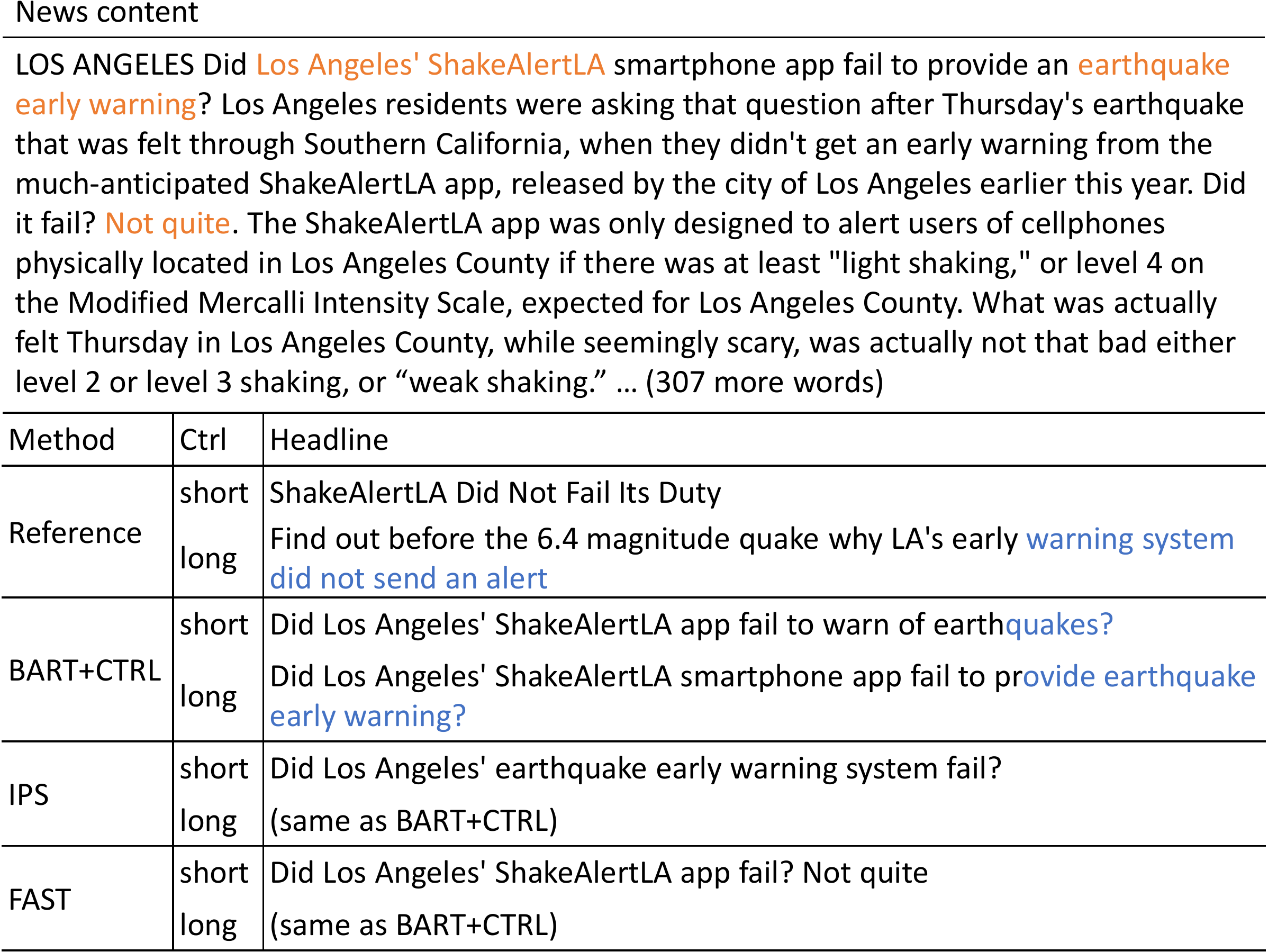}
\caption{An example from PENS dataset for news headline generation. Reference headlines from the balanced test set and generated headlines from three methods corresponding to the same news content are shown. Key news information that the generated headlines captured is highlighted in orange. Parts of the headline exceeding 55 characters are highlighted in blue.}
\label{fig:PENS_example}
\end{figure}

\subsection{Analysis}\label{sec:no_feedback_ablation}
\paragraph{Qualitative evaluation} Figure \ref{fig:PENS_example} shows example news headlines generated from three methods. While the BART+CTRL baseline failed to generate a short enough headline, generations from IPS and FAST fit within the 55 character length limit with FAST being the most concise. %For search ads generation, Figure \ref{fig:Ads_example} shows examples from BART+CTRL baseline and illustrate our motivation for controlling generation.
More comparative input-output examples for MReD and Ads datasets can be found in the Appendix.

% \begin{figure}[t]
% \includegraphics[width=20em]{ads_illustration.pdf}
% \caption{Illustration of controllable ads generation task. Each ad contains a title and a description as the arrows indicate. Two ads in different categories are generated for the same landing page. The second ad in the ``inventory'' category is more relevant for query ``used truck engine''. Key information in the landing page that the generated ads are based on is highlighted in orange.}
% \label{fig:Ads_example}
% \end{figure}

\paragraph{No Feedback ablation}
To see the importance of feedback in Step 3 of FAST algorithm (\S\ref{sec:fast}), we train a model without this step, i.e., self-training alone. Compared with FAST results in Table \ref{tbl:main_result}, the attribute/control-code accuracy drops by 9\%, 7\% and 12\% (absolute difference) on PENS, MReD and Ads respectively. On PENS, the control accuracy of self-training alone is even lower than the BART+CTRL baseline (73\% vs 78\%). In regular FAST training, 38\%, 25\% and 40\% of generated counterfactual samples are filtered out in the feedback step for PENS, MReD and Ads. It is not a surprise that including such a high percentage of noisy samples, whose attributes contradict with the control codes, would hurt controllability.

\paragraph{Classifier accuracy ablation}
We further study how the generation performance depends on the classifier accuracy. We replace the strong classifiers (Roberta-base for MReD and BERT-base-uncased for Search Ads) with weaker classifiers: l2-regularized logistic regression using bag-of-words features. On the hold-out test sets, the macro-F1 of the classifier drops from 79\% to 66\% for MReD and from 70\% to 58\% for Search Ads. In the feedback step, we filter out noisy examples using these weak classifiers and retrain the FAST models. Note that the controllability is evaluated in the same way with the strong classifiers. As shown in Table \ref{tbl:main_result}, the control accuracy of the ablation experiment (FAST with weak classifiers) drops from regular FAST, but it is still much higher than the BART+CTRL baseline. On the other hand, there is no statistically significant difference in ROUGE scores between ablation and regular FAST. This demonstrates that the improvement of FAST over baselines is fairly robust to the classifier accuracy.

% \begin{table}[t]
% \centering
% \resizebox{20em}{!}{
% {%
% \begin{tabular}{l|lll}
% \hline
% Acc                         & PENS          & MReD          & Ads \\ \hline
% BART+CTRL                   & 78            & 76.5          & 68.1\\ \hline
% FAST                        & \textbf{82.5} & \textbf{87.1} & \textbf{80.5}\\
% \hspace{0.5em} w/o feedback & 73            & 80            & 68 \\ \hline
% \end{tabular}%
% }
% }
% \caption{FAST feedback ablation experiment controllability results, compared to those in Table~\ref{tbl:main_result}. Removing the feedback in FAST training reduces the resulting controllability below the baseline level on the PENS dataset.}
% \label{tbl:feedback_ablation}
% \end{table}

\paragraph{IPS sampling mechanism}\label{sec:IPS_sub_sampling}
In Table \ref{tbl:main_result} we observed that IPS can reduce decoding quality (ROUGE score). We hypothesize that this may be because our procedure resamples the data with replacement, and the duplication of training samples can lead NLG models to memorize instead of generalizing \cite{feng2021survey}. To reduce the spurious correlation while not duplicating examples, we subsample 10k of training set according to IPS without replacement so each training sample is unique. For a fair comparison, we uniformly randomly subsample 10k for BART+CTRL training as the baseline. The results for the IPS subsampling and BART+CTRL are shown in Table \ref{tbl:IPS_ablation}. Now ROUGE scores between the two are close with no statistically significant difference. In addition, IPS improves controllability by 2.3\% from 71.9\% accuracy to 74.2\%, which is greater than its effect in oversampling experiments (1\% improvement from 78\% to 79\%). This demonstrates that rare examples are indeed more useful, especially if they are not duplicated.

\begin{table}[t]
\resizebox{20em}{!}{
\centering
{%

\begin{tabular}{l|lll|l}
\hline
Method   & R1            & R2           & RL            & Acc     \\
\hline
BART+CTRL & 31.4±0.2  & 12.7±0.2 & 26.3±0.2  & 71.9±1.1 \\
IPS       & 31.4±0.2  & 12.6±0.1 & 26.3±0.1  & \textbf{74.2}±0.5 \\
\hline
\end{tabular}

}
}
\caption{IPS subsampling ablation on PENS dataset. Best score with statistical significance is boldfaced.}
\label{tbl:IPS_ablation}
\end{table}

\section{Related Work}
\paragraph{Controllable generation} There are two main approaches for controllable generation: training or decoding-time steering. In the first approach, a language model is trained conditioned on the target attribute \citep{Ficler2017}, which can be conveniently encoded as control codes \citep{Keskar2019}. This approach has been used in many conditional generation tasks for controlling the length or content of abstractive summarization \citep{Kikuchi2016,Fan2018,Liu2018}, style of dialog response \citep{See2019}, ending of a story (happy or sad, conditioning on previous part of the story) \citep{Peng2018}, politeness of translation \citep{Sennrich2016}, intent of meta review (conditioning on individual reviews and ratings) \citep{Shen2021MReD}. As the training sets are usually collected through observation rather than intervention, we anticipate there exist shared confounders influencing both context and target attribute, causing the spurious correlation to be a pervasive problem. In the second approach, recent methods are PPLM \citep{Dathathri2020}, GeDi \citep{Krause2021}, FUDGE \citep{Yang2021} and DExperts \citep{Yang2021}. However, these methods exert control at the expense of fluency, a problem improved by \citet{Gu2022} but not completely eliminated. In contrast, our FAST method does not suffer from this problem.

\paragraph{Causal inference} Our IPS resampling method is motivated by causal inference. \citet{Feder2021} review causal inference in natural language processing and suggest to use causal knowledge to formalize spurious correlations and to mitigate predictor reliance on them. \citet{Hu2021} devise a structural causal model (SCM) for controllable generation, where the output text is the outcome and the attribute under control is the treatment (whether to write a short or long headline). To proceed with causal inference, there is a common challenge that observational (training) data is under selection bias as the treatment choice is affected by some confounders (context). A classical solution is IPS reweighting or resampling \citep{Yao2021,Pearl2009}. \citet{An2021} suggests that resampling works better with stochastic gradient descents than reweighting so we choose resampling to investigate primarily. Outside traditional causal inference areas, IPS reweighting is successfully applied in search ranking \citep{Wang2016} and recommendation systems \citep{Schnabel2016}. Recently, it is applied to reduce social bias in text classification tasks \citep{Han2021}. \citet{Hu2021} also propose to use it to debias pretrained language models.

\paragraph{Counterfactual data augmentation (CAD)} Our method FAST is a special case of CAD. \citet{Lu2018} propose CAD and generate synthetic examples to reduce the spurious correlations between gendered and gender-neutral words in training corpus. Similar rule-based techniques are most common for CAD. \citet{Zhao2018} build rules with crowd-sourced annotation to swap all male entities for female entities; \citet{Sharma2021} swap gender terms with a dictionary similar to \citet{Lu2018}; \citet{Garg2019} swap identity terms (e.g., gay, straight). Counterfactual examples can also be generated by manual post editing \citep{Kaushik2020,Gardner2020} or automated text rewriting \citep{Zmigrod2019,Riley2021,Wu2021}. Using the same model to augment data (self-training) is a common semi-supervised algorithm for improving classifier accuracy \citep{Zhang2022}, though less common for CAD. Most similar to our method FAST is \citet{Gu2019} for improving zero-shot neural machine translation via reducing the spurious correlation between the language of the output and the source sentence. They first train a model on the original data and use it to generate data in missing language pairs. A key difference is that our FAST method uses feedback, which is shown to be crucial in our scenario possibly because our spurious correlation is less severe.

\section{Conclusion}
This paper argues that conditional and controllable text generation systems are subject to spurious correlations in their training data which can severely undermine performance. We proposed a pair of simple yet effective data augmentation algorithms for countering this issue. One algorithm works by resampling the data according to an inverse propensity score, and the other via feedback-aware self training. Our experiments demonstrate that the proposed algorithms can effectively reduce the spurious correlation issue across three tasks: generating ad copy, news headlines, and meta-reviews. Furthermore these algorithms can significantly improve generation quality and controllability over popular and state-of-the-art baseline algorithms. 

Further research may investigate more checks during the feedback step, e.g., filtering out unfaithful examples. In the emerging parameter efficient fine-tuning paradigm, such as P*-tuning \citep{Li2021,Qian2022}, we find IPS resampling to be promising as the model is not likely to memorize duplicate examples when updating only few parameters. The proposed method may also be complementary to baselines like GeDi and PPLM.

\section{Limitations}
While IPS is a classical technique from causal inference to deal with spurious correlations, we found the following limitations when applying it to controllable text generation, which makes it less effective than FAST. First, \citet{Tu2020} found that large pretrained models are quite efficient in learning small amounts of counterfactual examples, which makes them more robust to spurious correlations. Our IPS resampling makes the small amounts of counterfactual examples more important to learn, but may have limited impact on the large pretrained models. Second, for MReD and Search Ads, the human-labeled or classifier-detected categories could be wrong. These examples are likely to have low propensity scores and therefore get up-sampled by IPS method. Finally, some unique training examples are dropped after resampling.

For FAST method, we acknowledge two limitations. First, an implicit assumption is that the linguistic attribute of interest (headline being short or long) should be independent of the context, therefore, a control code is applicable for any context. We design ad categories with this consideration in mind. However, in MReD dataset, categories such as ``rebuttal process'' may not be applicable for every meta review. Forcing a model to produce a sentence in such categories may result in untruthful generations. Second, FAST may struggle if the training and pre-training data are drastically different; the counterfactual generations may be of low quality and propagate errors into the final FAST model.  How to generate counterfactual data in those more challenging scenarios would be our future research direction.

\bibliography{anthology,custom}
\bibliographystyle{acl_natbib}

\appendix

\section{Appendix}

\subsection{Hyperparameters}
Hyperparameters in use are summarized in in Table \ref{tbl:hyper_params}. We use the default HuggingFace setup unless otherwise specified (e.g., default learning rate scheduler, default length and repetition penalties). We tune learning rate from \{1e-5, 5e-5, 1e-4\} and pick the ones with the best validation metric. Number of epochs are reported for BART+CTRL baseline training. Uncontrolled, IPS, FAST, and GeDi are trained for similar epochs. Generation speed is reported for BART+CTRL baseline. The maximum number of tokens are picked so that 99\% of the times input/ouput from training set will not get truncated, except if it exceeds the maximum number of tokens of the pretrained models.

\begin{table}[t]
\centering
\resizebox{20em}{!}{
{%
\begin{tabular}{lllll}
\hline
model                       & params                           & PENS     & MReD     & Ads      \\
\hline
generation & batch size                       & 64       & 36       & 96       \\
                            & \# epochs                        & 40       & 10       & 6        \\
                            & LR                               & 1e-5 & 5e-5 & 5e-5 \\
                            & encoder \#tokens                 & 768      & 1024     & 314      \\
                            & decoder \#tokens                 & 32       & 72       & 26       \\
                            & train speed      & 70       & 25       & 190      \\
                            & generation speed & 85       & 40       & 150      \\
\hline
propensity & batch size                       & 64       & 64       & 64       \\
                            & input \#tokens                   & 512      & 512      & 314      \\
                            & \# epochs                        & 6        & 15       & 5        \\
                            & LR                               & 1e-5 & 1e-5 & 1e-5 \\
                            & train speed      & 70       & 25       & 110      \\
                            & inference speed  & 240      & 170      & 470     \\
\hline
\end{tabular}%
}
}
\caption{Hyperparameter settings for generation model and propensity model on PENS and Ads datasets. Unit for training/inference speed is number of samples per second on one GPU.}
\label{tbl:hyper_params}
\end{table}

\subsection{Propensity model}
We finetune Roberta-base for sequence classification with the default HuggingFace setup, to predict the attribute of the ground truth target from the context as input. On PENS, we do a binary classification; therefore we use a sigmoid function to transform the score into probability. On MReD and Ads, we do multi-class classification task and use a softmax fucntion to convert scores into probabilities. We also use the dev set to pick the best epoch during training based on AUC for PENS and accuracy for MReD and Ads.

\subsection{PPLM}
We follow the implementation from \citet{Dathathri2020} with minimal changes to adapt it to an encoder-decoder structure (BART). We apply perturbations on both cross-attention and self-attention key-value pairs instead of only the self-attention as in the original paper for decoder-only models (GPT2). PPLM uses a discriminative classifier $p(a|y)$ to steer decoding towards having the desired attribute. In addition, the classifier needs to have the same vocabulary as the generation model. Therefore, we finetune Roberta-base for this purpose as it shares the same vocabulary with BART. The Roberta-base classifiers achieve 96\% and 88\% macro-F1 for PENS and Ads respectively on the default test sets. Note that on Ads data, we use a previously built BERT-base classifier to produce attribute label on the training and testing data (Table \ref{tbl:summary}). This Roberta-base classifier is trained and tested on the labels predicted by the previous BERT-base model, thus mimicking its behavior. For MReD, we reuse the Roberta-base classifier previously described in \S\ref{sec:dataset}.

To steer decoding, PPLM ascends $p(a|y)$ by propagating gradients from the classifier to the perturbations on key-value pairs in BART. At each decoding step, BART outputs a probability distribution $p(y_i|y_{<i}, ...)$ for generating the current $i$th token. We feed the classifier with previous $i-1$ tokens plus a ``soft'' token for $i$, which is a weighted average of embeddings $\sum_{y_i\in V} p(y_i|y_{<i}, ...) E(y_i)$, where $V$ denotes the vocabulary and $E(\cdot)$ denotes the input embedding of the classifier (Roberta). Therefore, we can propagate gradients from the classifier into BART.

We tune hyperparameters on dev set to find a good balance between controllability and decoding quality. We tune $\gamma_{gm}$ from \{0.1, 0.3, 0.5, 0.8\}, $\lambda_{KL}$ from \{0, 0.005, 0.01\}, and we settle with 0.3 and 0.01 for them respectively. As PPLM decoding output tends to be repetitive, we additionally tune repetition penalty to be 2.

We use greedy decoding following the original implementation. Using beam search for PPLM would be computationally prohibitive as it is already very slow.

\subsection{GeDi}
We train BART-base for the two models in GeDi. One model $p(y|x)$ uses context $x$ as the encoder input, and the other model $p(y|c)$ uses control code $c$ as the encoder input. The hyperparameters are the same as BART+CTRL baseline except that the max number of encoder tokens for $p(y|c)$ model is 3, 7, and 8 on PENS, MReD, and Ads datasets respectively. We use negative log likelihood (instead of ROUGE1) as the validation metric during training $p(y|c)$. Same as the original GeDi, an additional length normalization heuristic is applied in the Bayes rule for computing the steering term during decoding:
\begin{equation}
p(c|y) = \frac{p(c)p(y|c)^{1/t}}{\sum_{c'\in \{1,...,K\}}{p(c')p(y|c')^{1/t}}},\label{eq:Gedi_contrast_weight}
\end{equation}
where $t$ is the current length of $y$. We avoid tuning hyperparameters with the following choices. We choose uniform prior for $p(c)$, and we do not apply the filtering heuristic during decoding but we use the standard beam search as the other methods. We train $p(y|c)$ as a usual generation model conditioned on control code $c$. In the end, we only tune the parameter $\omega$ to control the trade-off between controllability and decoding quality.

The original GeDi paper converts multi-class classification into multiple binary classification tasks with control code and anti-control code for each class in order to improve speed. We do not do this conversion but normalize over all $K$ classes in Eq. \ref{eq:Gedi_contrast_weight}.

\subsection{GeDi+x}
We reuse our implementation of GeDi for GeDi+x by simply switching both $p(y|x)$ and $p(y|c)$ models to BART+CTRL baseline $p(y|x,c)$. So the settings are the same as GeDi or BART+CTRL.

\subsection{Computational cost}
We compare the training and inference cost for different methods on Search Ads dataset in Table \ref{tbl:computing_cost}. During training, FAST has the largest cost due to the additional cost from training the initial model and using it to generate counterfactual data. GeDi also has higher cost than the rest of the methods as it trains an additional model $p(y|c)$ for steering the decoding. During generation, PPLM is much slower than other methods at it needs to compute gradient several times at each decoding step. GeDi and GeDi+x are also slow due to computing the contrast term (e.g., Eq. \ref{eq:Gedi_x_contrast_weight}). The rest of the methods are equally fast.

\begin{table}[t]
\centering
\resizebox{20em}{!}{
{
\begin{tabular}{lll}
\hline
Method                 & Training (hrs) & Generation (samples/sec) \\
\hline
BART+CTRL & 30                    & 150                                 \\
PPLM                   & 30                    & 0.38                                \\
GeDi                   & 42                    & 8                                   \\
GeDi+x                 & 30                    & 8                                   \\
IPS                    & 30                    & 150                                 \\
FAST                   & 85                    & 150 \\ \hline                                
\end{tabular}%
}
}
\caption{Training time and generation speed for different methods measured on 1 Nvidia V100 GPU.}
\label{tbl:computing_cost}
\end{table}

\subsection{Spurious correlation in MReD}
Similar to PENS dataset, we finetune Roberta-base to predict the next sentence's category from the context (all preceding sentences, ratings from individual reviewers and their reviews). The classifier achieves 71\% AUC and 33\% accuracy on hold-out test set, which are much higher than random guess (50\% AUC, 11\% accuracy) or majority class (13\% accuracy). Therefore, we empirically confirm the existence of spurious correlation between context and MReD category.

Results in the original MReD paper have already implied some reasons for why such correlations exist. First, preceding sentences are predictive of the category of the next sentence because meta-reviews have some typical patterns, for example, ``abstract$\rightarrow$strength$\rightarrow$weakness'', ``rating summary$\rightarrow$weakness$\rightarrow$rebuttal process'', and etc. Second, individual reviewers' ratings and their reviews are predictive of the category of a sentence in the meta review. For example, there is higher chance to get a ``strength'' than ``weakness'' in the meta review if the individual reviewers are more positive about a paper.

To show the damaging effect of the spurious correlation, we use BART+CTRL baseline to generate next sentences in all 9 categories. Then we detect the category from output using the Roberta-base classifier. Finally, we evaluate the accuracy between the detected category and the intended category (control code). The system successfully generated the intended category 90\% of the time in factual cases (when the control code matched the ground-truth attribute), but 75\% of the time in counterfactual cases. Again, this demonstrate the degradation of controllability when spurious correlations break at test time.

% \subsection{Background on search ads generation}
% Our study is mainly motivated by its industrial application in automating search advertising. Search engines derive most of their revenue by displaying ads along with the search results. To start a traditional advertising campaign, advertisers need to manually create ads for their landing pages, which are the web pages to visit when users click the ads. The Dynamic Search Ads product automates this process: the advertisers can simply provide us their website domains to start a campaign. Our Web Indexing infrastructure crawls all the landing pages under their domains and our Document Understanding pipeline parses the landing page HTMLs to extract textual features such as document title and heading. We use various techniques including CLMs to generate ads from landing page features, which are then ingested into the online data store. Finally, the online ranking and auction systems decide which ads to display in response to user query.

\subsection{Spurious correlation in Search Ads}
Similar to PENS and MReD datasets, we finetune Roberta-base to predict the ad category from the context, which include various landing page features. The classifier achieves 74\% AUC and 36\% accuracy on hold-out test set, which are much higher than random guess (50\% AUC, 11\% accuracy) or majority class (23\% accuracy). Therefore, we empirically confirm the existence of spurious correlation between context and ad category.

This correlation is hardly a surprise. Advertisers write ads that perform well for their landing pages on average, so different categories are preferred for different landing pages. While the majority category is product or service on all data as shown in Table \ref{tbl:summary}, by slicing data into different business industries, we find that majority category is location for travel and tourism industry, call to action for vehicle industry, and highlight for retail industry (which contains promotion, shipping or other information to make the product stand out). While an ad in the majority category may perform well on average, we can get an even better chance to win the user click by generating ads in all categories and displaying the best one at query time. For example, while ``Buy Truck Engines Now'' may be a good ad for query ``truck engine'', ``New \& Used Truck Engines'' is a better choice for query ``used truck engine''.

We then use BART+CTRL baseline to generate ads in all 9 categories. Then we detect the category from output using the BERT-base-uncased classifier. Finally, we evaluate the accuracy between the detected category and the intended category (control code). The system successfully generated the intended category 76\% of the time in factual cases (when the control code matched the ground-truth attribute), but 65\% of the time in counterfactual cases. Again, this demonstrate the degradation of controllability when spurious correlations break at test time.

\subsection{Details on human evaluation}
As opposed to crowd sourced judges, our judges are paid with hourly wages and they are doing the labeling task for a long term, therefore they demonstrate more consistent labeling quality. They have been trained to ensure understanding the tasks correctly and they get feedback from us to ensure their labeling quality and consistency.

The quality of an ad is labelled in 4 aspects: 1) language, which checks spelling/capitalization, grammar, fluency; 2) human-like, which checks if the ad sounds like human written rather than machine generated and if it agrees with common sense; 3) factuality, which checks if the ad contains false claims (e.g., free shipping) not existing in the  landing page; and 4) relevance, which checks if the ad is relevant for the landing page.  Judges should visit the landing page and read it carefully for checking factuality and relevance. Judges can also skip an example in cases such as the text is in a foreign language or the landing page is not accessible.

For category labeling, judges first select if an ad is scorable to prevent cases such as text is in a foreign language or quality is too poor to understand. Judges are trained before they start labeling by going through our judgement guideline and passing our test judgement task. In our judgement guideline, we explain definition of each category with examples. We also explain the idea behind designing these categories that advertising is commonly surrounding three roles -- advertisers (their name or brand), products (what's the product, purchasing information such as price, shipping), and customers (what's the benefit for customer, call to action) -- to help judges better differentiate these categories.

\subsection{Examples of generated meta reviews}
\begin{figure*}
\includegraphics[width=\textwidth]{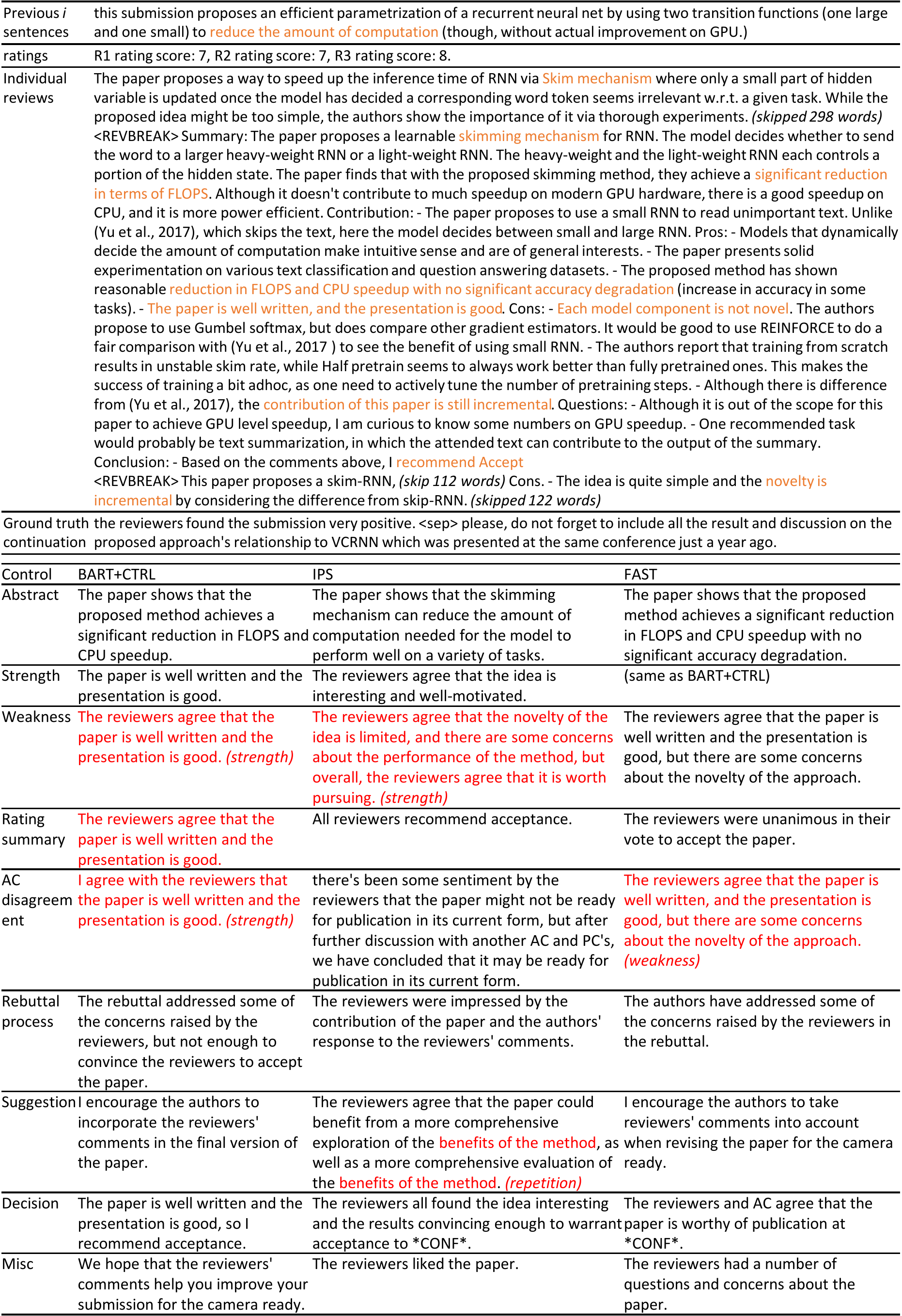}
\caption{Example of generated next sentence for a meta review. Key information in the input that the generations are based on is highlighted in orange. Issues with generations are highlighted in red with a brief explanation.}
\label{fig:MReD_examples}
\end{figure*}

We provide an example of generations from 3 models (BART+CTRL baseline, IPS and FAST) in Figure \ref{fig:MReD_examples}. In this example, the individual reviews are quite positive, and so is the ground truth meta review. The BART+CTRL model seems to struggle in generating a sentence in the ``weakness'' category, but it is preferring ``strength'', thereby ignoring the control code, which is also the case for ``rating summary''. On the other hand, FAST is able to generate a ``weakness'' sentence correctly. For ``AC disagreement'' category, even FAST struggles to generate it correctly. This is likely due to the fact that there are only 24 training examples in this category. Interestingly, IPS generates a correct example in this category, which seems to the case in general as we examined more examples. However, IPS suffers from worse language quality. It generates a sentence with repetition issue in the ``suggestion'' category. We note that the ``rebuttal process'' category should not be applicable for this meta review, as there is no such information from the context.

\subsection{Examples of generated ads}
\begin{figure*}
\includegraphics[width=\textwidth]{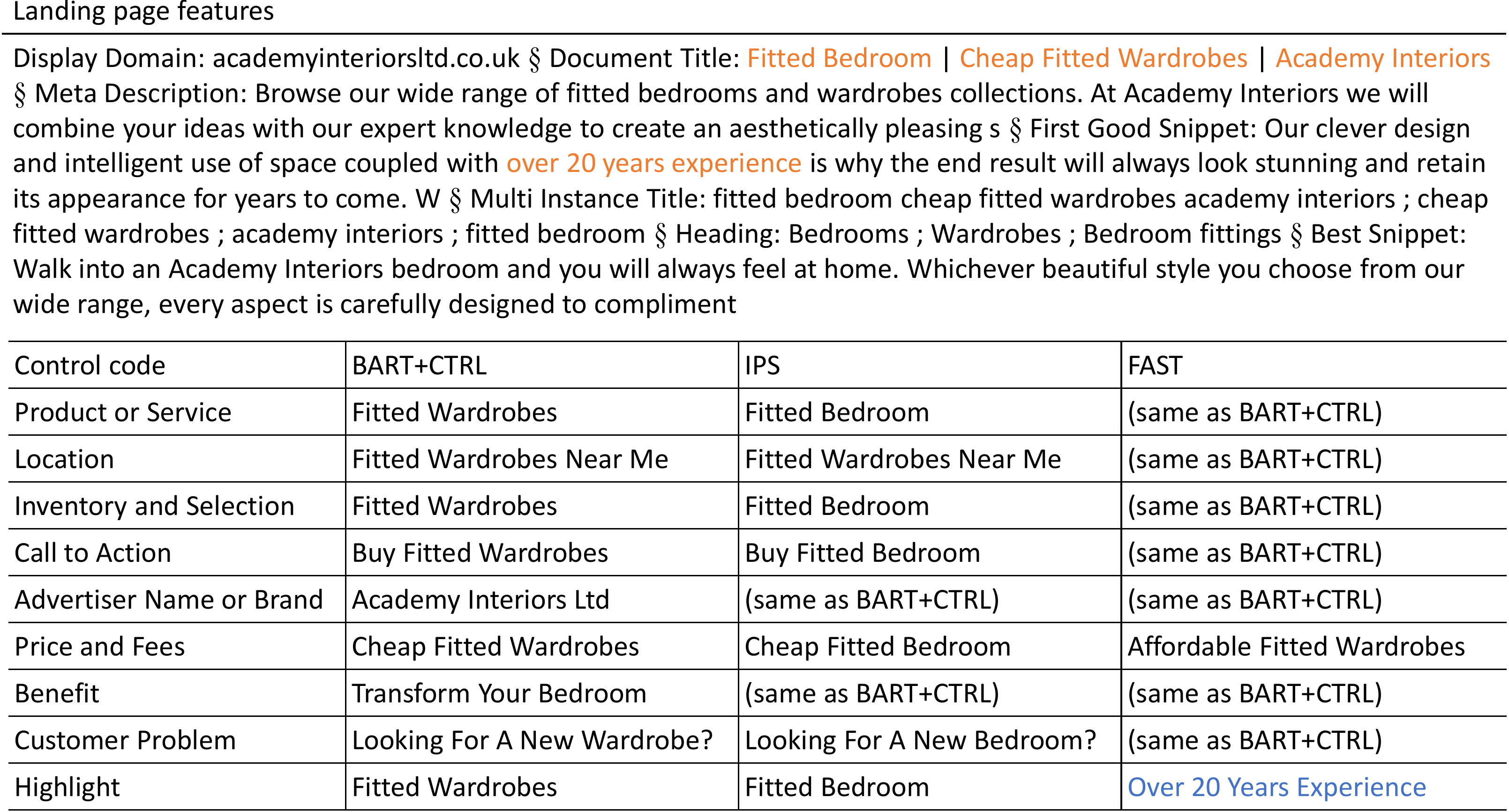}
\caption{Example of generated ad titles. For this landing page, only FAST generates title correctly in ``Highlight'' category. Key information in the landing page that the generated ads are based on is highlighted in orange.}
\label{fig:Ad_title_examples}
\end{figure*}

\begin{figure*}
\includegraphics[width=\textwidth]{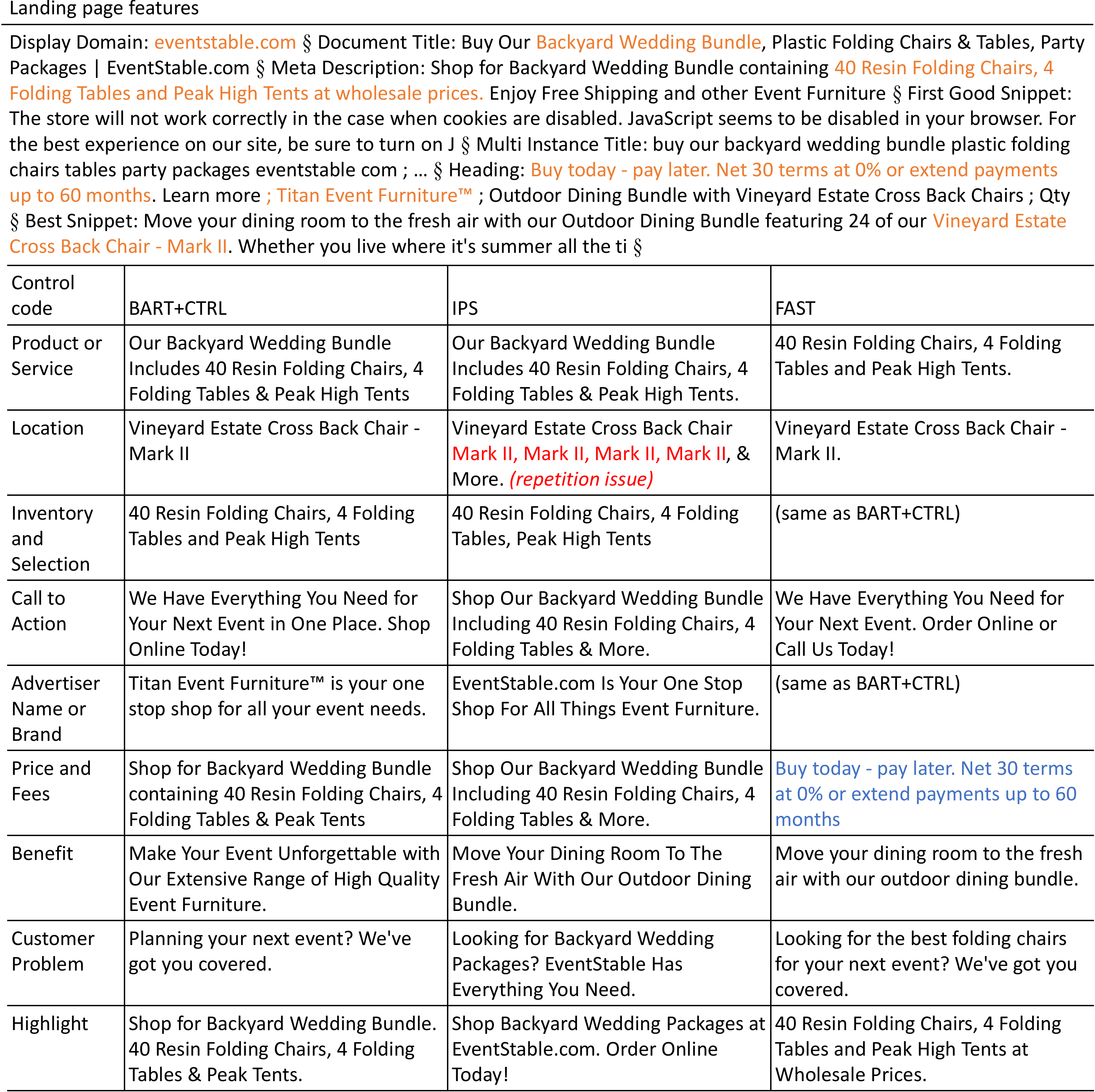}
\caption{Example of generated ad descriptions. For this landing page, only FAST generates description correctly in ``Price and Fees'' category. One description from IPS model has repetition issue. Key information in the landing page that the generated ads are based on is highlighted in orange}
\label{fig:Ad_desp_examples}
\end{figure*}

We provide examples of generated ads from 3 models (BART+CTRL baseline, IPS and FAST) in Figure \ref{fig:Ad_title_examples} and \ref{fig:Ad_desp_examples}. As the difference between the 3 models are not huge as seen from the human evaluation results (although statistically significant), we pick those examples to highlight the typical difference between the 3 models. In actual online serving, up to 3 titles can be concatenated together with delimiter | to form a longer title, and up to 2 descriptions can be concatenated together.

\end{document}